\documentclass[preprint,12pt,authoryear]{elsarticle}

\usepackage{graphicx}
\usepackage{amssymb}

\usepackage{lineno}

\usepackage{url}
\usepackage{xcolor}
\usepackage{multirow}
\usepackage{bigstrut}

\usepackage{adjustbox}
\usepackage{amsmath}

\usepackage{multicol}
\usepackage{tabularx}

\usepackage{hyperref}
\usepackage{graphics}

\DeclareMathOperator{\similarity}{sim}
\DeclareMathOperator*{\argmax}{argmax}

\journal{Expert Systems with Applications}

\begin{document}

\begin{frontmatter}


\title{W2VLDA: Almost Unsupervised System \\ 
for Aspect Based Sentiment Analysis}





\cortext[cor1]{Corresponding author}

\author[vic]{Aitor Garc\'ia-Pablos\corref{cor1}}
\ead{agarciap@vicomtech.org}
\author[vic]{Montse Cuadros}
\ead{mcuadros@vicomtech.org}
\author[ehu]{German Rigau}
\ead{german.rigau@ehu.eus}
\address[vic]{Vicomtech-IK4, Mikeletegi 57, San Sebastian, Spain}
\address[ehu]{IXA Group, EHU, Manuel Lardizabal 1, San Sebastian, Spain}

\begin{abstract}

With the increase of online customer opinions in specialised websites and social networks, the necessity of automatic systems to help to organise and classify customer reviews by domain-specific aspect/categories and sentiment polarity is more important than ever.
Supervised approaches for Aspect Based Sentiment Analysis obtain good results for the domain/language they are trained on, but having manually labelled data for training supervised systems for all domains and languages is usually very costly and time consuming. In this work we describe W2VLDA, an almost unsupervised system based on topic modelling, that combined with some other unsupervised methods and a minimal configuration, performs aspect/category classification, aspect-terms/opinion-words separation and sentiment polarity classification for any given domain and language. We evaluate the performance of the aspect and sentiment classification in the multilingual SemEval 2016 task 5 (ABSA) dataset. We show competitive results for several languages (English, Spanish, French and Dutch) and domains (hotels, restaurants, electronic devices).
\end{abstract}


\end{frontmatter}


\section{Introduction}
\label{sec:intro}

During the last decade, the Web has become one of the most important sources for customers and providers to evaluate and compare products and services. The vast amount of content generated every day in countless websites and social networks keeps growing and requires automated ways to handle and classify all these opinions. Because of that, many different algorithms and approaches have been developed in the area of Opinion Mining.

\begin{figure*}[t] 
	\centering
	\includegraphics[width=1.0\textwidth]{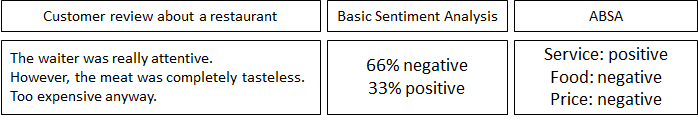}
	\caption{An example of classical Sentiment Analysis vs. Aspect Based Sentiment Analysis}
	\label{figure:example_sentence}
	
\end{figure*}

Opinion Mining is a subfield of Natural Language Processing (NLP) that deals with the automatic analysis of opinions shared by humans in different contexts, like in customer reviews \citep{pang2008opinion,liu2012sentiment}.
Aspect Based Sentiment Analysis (ABSA) refers to the systems that determine the opinions or sentiments expressed on different features or aspects of the products/services under evaluation (e.g. “battery” or “performance” for a laptop).
An ABSA system should be capable of classifying each opinion according to the aspects relevant for each domain in addition to classifying its sentiment polarity (usually positive, negative or neutral), as depicted in figure \ref{figure:example_sentence}.

Best performing ABSA systems generally use manually labelled data and language specific resources for training on a particular domain and for a particular language \citep{Pontiki2014,pontiki2015semeval,pontiki-EtAl:2016:SemEval}. This is the case of deep-learning based systems, that provide very good performance but require a significant amount of labelled data for training \citep{chen2017improving,araque2017enhancing}.

On the other hand, weakly-supervised systems do not require labelled data for training, but they usually need some language specific resources, such as carefully curated lists of seed words or language dependent tools to preprocess the input 
\citep{Lin2012,Jo2011,KimSuin2013}.
In addition, most of these works only report results for English. 

\begin{figure*}[t!] 
	\centering
	\includegraphics[width=1.0\textwidth]{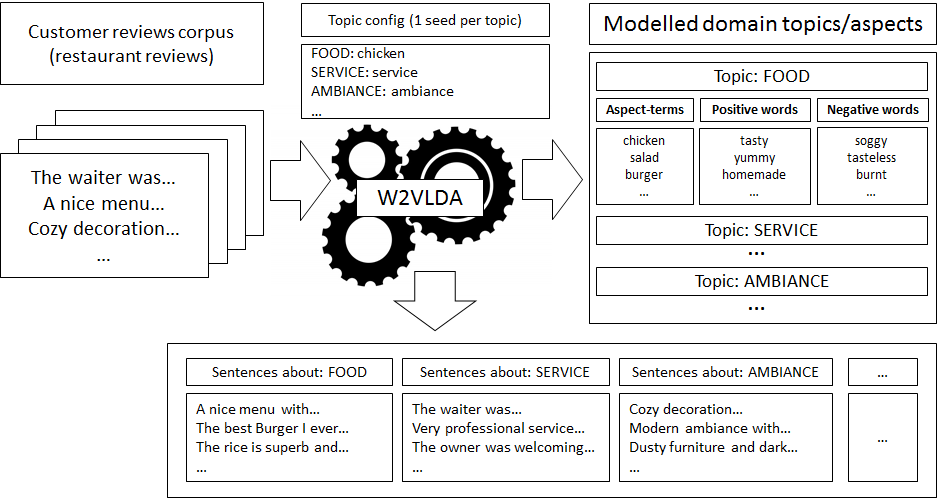}
	\caption{An schema of W2VLDA. The input is an unlabelled corpus o a particular domain and the topic specification. Topics are split into three word distributions: aspect-terms, positive words and negative words to ease the interpretation of each topic. Sentences are modelled by topic/aspect and polarity.}
	\label{figure:example_modelling_work}
	
\end{figure*}

In this work, we present W2VLDA, an almost unsupervised system for multilingual and multidomain ABSA, that works leveraging large quantities of unlabelled textual data and an initial configuration consisting of a minimal set of seed words. Figure \ref{figure:example_modelling_work} shows an schema of W2VLDA. 
Imagine the following scenario. The owners of a famous restaurant want to monitor the opinion of their costumers with respect to a set of aspects. In particular, they want to know the opinion about its food, service, price, ambience, location, etc. The input of W2VLDA is a corpus of customer reviews and an example word per aspect they want to monitor (for instance, chicken for the aspect food, service for the aspect service, etc.)\footnote{W2VLDA also needs an example of a positive and a negative word (for instance, excellent and horrible).}. With this input, W2VLDA produces two main outputs. First,  a weighted list of words per aspect (for instance, chicken, salad, burger, etc. for the aspect food), a weighted list of positive words (tasty, yummy, homemade, etc.) and weighted list of negative words (soggy, tasteless, burnt, etc.) for every selected aspect. Thus, our system performs at a word level three subtasks simultaneously: aspect classification, aspect-term/opinion-word separation, and sentiment polarity classification. Second, W2VLDA also produces a weighted list of sentences for every selected domain aspect and polarity.

The system is based on a topic modelling approach combined with continuous word embeddings and a Maximum Entropy classifier. It runs over an unlabelled corpus of the target language and domain just by defining the desired aspects with a single seed-word per aspect.
We show results for different domains (restaurants, hotels, electronic devices) and languages (English, Spanish, French and Dutch). We compare its performance with other topic modelling based approaches, and we evaluate the performance of this approach on the SemEval2016 task 5 dataset, which provides a manually labelled set of restaurant reviews for several languages, including English, Spanish, French and Dutch.
The contributions of this work are the minimal need of supervision (just one seed word per aspect/polarity) to perform ABSA over any unlabelled corpus of customer reviews. The lack of language or domain specific requirements allows the system to be readily used for other languages and domains. Another contribution is the automatic separation of the topic words into aspect-terms, positive words and negative words to improve the readability of the generated topics.
We will leave the source code publicly available\footnote{\url{https://bitbucket.org/aitor-garcia-p/w2vlda-last/overview}}.

After this short introduction, the paper is structured as follows. First, section \ref{sec:related_work} reviews previous related work. Then, section \ref{sec:approach} describes our system, including the seed-word based configuration, the aspect-term/opinion-word separation and the topic modelling part. After that, section \ref{sec:results} shows the results and evaluation. Finally, section \ref{sec:conclusions} describes the conclusions and future work.

\section{Related work}
\label{sec:related_work}

During the last decade the research community has addressed the problem of analysing user opinions, particularly focused on online customer reviews \citep{Liu2012,Liu2014a}.
The problem of customer opinion analysis can be divided into several subtasks, such as detecting the aspect (aspect classification) and detecting the opinion about the aspect of the product being evaluated.

A common approach in the literature is to identify frequent nouns, lexical patterns, dependency relations applying supervised machine learning approaches \citep{Hu2004,Popescu2005,Blair-Goldensohn2008,wu2009phrase,Qiu2011}. Some works focus on automatically deriving the most likely polarity for words, constructing a so-called sentiment lexicon \citep{mostafa2013more}. The typical approaches use different variants of bootstrapping or polarity propagation leveraging some base dictionaries and pre-existing linguistic resources \citep{rao2009semi,jijkoun2010generating,huang2014automatic}.

A well-known unsupervised method for text modelling documents is Latent Dirichlet Allocation (LDA). LDA is a generative model introduced by \citep{blei2003latent} that quickly gained popularity because it is an unsupervised, flexible and extensible technique to model documents. LDA models documents as multinomial distributions of so-called topics. Topics are multinomial distributions of words over a fixed vocabulary. Topics can be interpreted as the categories from which each document is built up, and they can be used for several kinds of tasks, like dimensionality reduction or unsupervised clustering. Due to its flexibility, LDA has been extended and combined with other approaches, obtaining topic models that improve the resulting topics or that model additional information \citep{mcauliffe2008supervised,ramage2009labeled}.

Topic models have been applied to Sentiment Analysis to jointly model topics and sentiment of words \citep{Lin2009,Lin2012,Jo2011,Lu2011,KimSuin2013,Alam2016}. A usual way to guide a topic modelling process towards a particular objective is to bias the LDA hyperparameters using certain apriori information. In the case of modelling the polarity of the documents, it usually means using a carefully selected set of seed words. Our method follows this idea, but replaces the need for a carefully crafted list of language or domain polarity words by only a single domain independent positive word (e.g. \textit{excellent} for English) and a single domain independent negative word (e.g. \textit{horrible} for English).

In general, topics coming from a topic modelling approach are anonymous word distributions, requiring an additional step to map them to a meaningful domain category. This task requires a manual inspection by an expert or a mapping calculation to an existing resource \citep{bhatia2016automatic}. Our approach relies on a minimal topic configuration to define the topics for the target domain the user wants to monitor. Thus, the resulting topics match the ones defined initially by the user. This is done by leveraging semantic word similarities to guide the topic modelling towards the defined topics. This semantic word similarity is obtained using continuous word embeddings over the domain words. Continuous word embeddings are known for capturing semantic regularities of words \citep{Mikolov2013,collobert2008unified}. Some works have made use of this fact to improve the resulting topics \citep{das2015gaussian,Nguyen2015,Qiang2016}, but their objective is to improve the unsupervised modelling of a corpus instead of guiding the model towards a predefined set of topics. There are works that exploit word embeddings in a supervised machine learning setting to perform sentiment analysis \citep{Tang2014,giatsoglou2017sentiment}.

Some authors have also attempted an automatic aspect-term/opinion-word separation within the topic modelling process \citep{Zhao2010,Mukherjee2012}. Aspect terms are the words that are used to speak about the aspect being evaluated (e.g. \textit{waiter} or \textit{waitstaff} when speaking about the \textit{service} of a restaurant). On the other hand, opinion words express the sentiment about an aspect, such as \textit{attentive} or \textit{terrible}. The separation of these two kinds of words might be useful because it eases the interpretation of the resulting topics, and the sentiment classification can be focused on the opinion-words which are more likely to bear sentiment information. \citet{Zhao2010} attempted this separation training a supervised classifier on a small manually labelled dataset and using Part-of-Speech tagging. \citet{Mukherjee2012} elaborated on this idea trying a similar approach but substituting the manually labelled dataset with an existing lexicon of opinion words for English. Instead, we apply Brown clustering \citep{Brown1992} to a set of training instances from an unlabelled corpus in order to train an aspect-term/opinion-word classifier that is later integrated into the topic modelling process. Following this approach, no additional language-dependent resources are required, and the full process could be applied to any language and domain.

In summary, combining topic modelling, continuous word embeddings and a minimal topic definition, our proposed approach can model customer reviews in different languages and domains performing three subtasks at the same time: aspect classification, sentiment classification and aspect-terms/opinion-words separation. To our knowledge, no other almost unsupervised system tries to perform these three tasks at the same time and without requiring any pre-existing language or domain dependent resource.

\section{System description}
\label{sec:approach}

The main objective of the W2VLDA system is to perform the three tasks  (detecting aspects, opinions and their polarity) of Aspect Based Sentiment Analysis at the same time. That is, to classify pieces of text into a predefined set of domain aspects and classify their sentiment polarity as positive or negative. 
In addition, our system separates opinion words from aspect terms without requiring additional resources or supervision. The system at its core consists of an LDA-based topic model extended with additional variables, with biased topic modelling hyperparameters based on continuous word embeddings, and combined with unsupervised pre-trained classification model for aspect-term/opinion-word separation.

\subsection{Topics and sentiment configuration}

\begin{table}[t]
	\footnotesize
	\centering
	\begin{tabular}{|l|l|l|l|}
		\hline
		Domain aspect or Polarity & Seeds (English) & Seeds (Spanish) & Seeds (French) \bigstrut\\
		\hline
		food  & chicken & pollo & poulet \bigstrut\\
		\hline
		service & service & servicio & service \bigstrut\\
		\hline
		ambience & ambience & ambiente & ambiance \bigstrut\\
		\hline
		drinks & drinks & bebidas & boissons \bigstrut\\
		\hline
		location & location & ubicaci\'on & emplacement \bigstrut\\
		\hline
		\textit{positives} & excellent & excelente & excellent \bigstrut\\
		\hline
		\textit{negatives} & horrible & horrible & \'epouvantable \bigstrut\\
		\hline
	\end{tabular}%
	\caption{Example of seed words (one per domain aspect) used to monitor certain aspects of restaurant reviews in several languages, including the general polarity seeds}
	\label{tab:seeds_restaurants}%
\end{table}%

W2VLDA only requires a minimal domain aspect and sentiment polarity configuration per language and domain. The configuration consists of a single seed to define each desired domain aspect, plus a single general positive seed word and a single general negative seed word valid for all domain aspects. This simple configuration is the only language and domain dependent information required by W2VLDA \footnote{A list of general stopwords for each target language is also necessary in order to obtain better results. We use the stopword lists from Apache Lucene.}. Therefore, a simple translation of the seeds should suffice to make the system work for another language or domain, as long as each translated seed has an equivalent meaning and use in the target language. Table \ref{tab:seeds_restaurants} shows an example of a domain aspect and polarity configuration for the restaurant domain in several languages.

\subsection{Aspect-term and opinion-word separation}
\label{sec:atow_separation}

Part of the outcome of the system consists of the aspect-term/opinion-word separation into differentiated word classes. In order to achieve this separation without adding any language dependent tool or resource, the system uses Brown clusters \citep{Brown1992} to model examples of aspect-terms and opinion-words and train a MaxEnt-based classification model.
Brown clusters have been used as unsupervised features with good results in supervised Part-of-Speech tagging \citep{turian2010word} and Named Entity Recognition \citep{Agerri201663}.
Brown clusters are computed\footnote{We use the Brown clustering implementation at \url{https://github.com/koendeschacht/brown-cluster}} from the domain unlabelled corpus with no additional supervision, and are used as the features for the two words context window, [-2,+2], of each training example. The training instances are obtained leveraging the occurrences of the initial configuration with aspects and polarity seed words, assuming that domain aspect seed words are aspect-terms and polarity-words are opinion-words.

\begin{figure*}[t] 
	\centering
	\includegraphics[width=1.0\textwidth]{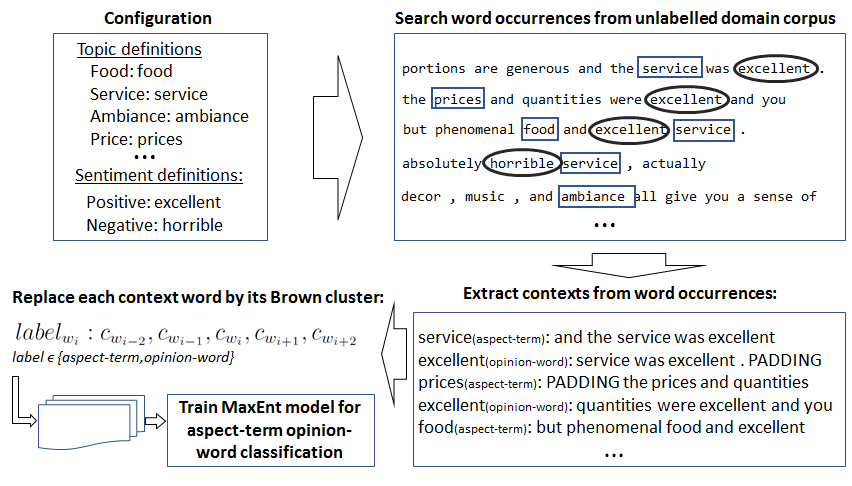}
	\caption{Process to obtain the MaxEnt model for aspect-term/opinion-word separation. 
	}
	\label{figure:brown_clusters_process}
\end{figure*}

Figure \ref{figure:brown_clusters_process} describes the process to obtain the classification model. First domain aspect seed words and polarity seed words are used as gold aspect-terms and gold opinion-words respectively. Then the occurrences of these words are bootstrapped from the domain corpus and they are modelled according to their context window. Next, context words are replaced by their corresponding Brown cluster to build each training instance. Finally, a MaxEnt model is trained using these generated training instances.

We have experimented with a different number of Brown clusters (100, 200, 500, 1000 and 2000) but the impact of this parameter was negligible for this purpose. The reported results have been obtained using 200 clusters.

A drawback of this approach is that every word in the vocabulary will be classified as aspect-term or as opinion-word. There are words that do not belong to any of these categories. It would be interesting to have a third class (e.g. \textit{"other"}), but it would require labelling training instances for that additional class, introducing a manual supervision that we want to keep to a minimum. We assume that the words that are not clearly aspect-terms or opinion-words will be spread across both classes, losing relevance during the topic modelling process.


\subsection{Combining everything in a topic model}
\label{sec:topic_model}

The core of the system consists of an LDA-based topic model, extended to include the aspect-term/opinion-word separation and the positive/negative separation for each topic. In addition, the aspect-term/opinion-word separation is guided by a pre-trained classifier as explained at section \ref{sec:atow_separation}, while the topic and polarity modelling are guided by biasing certain hyper-parameters according to the given topic configuration.

\begin{figure*}[t]
	
	\centering
	\includegraphics[width=1.0\textwidth]{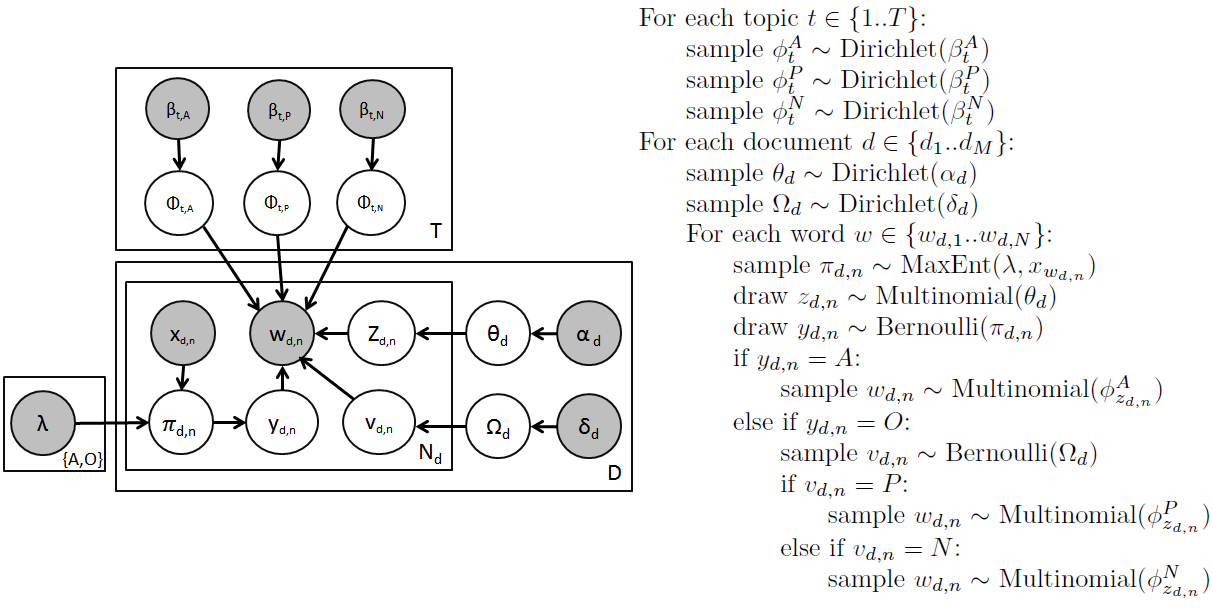}
	\caption{Proposed model in plate notation and its generative process algorithm.}
	\label{figure:process}
	
\end{figure*}

Figure \ref{figure:process} shows the proposed model in plate notation and the generative story modelled by the algorithm.

The generative hypothesis described by the model is the following. For each document $d$ a distribution of topics, $\theta_{d}$, is sampled from a Dirichlet distribution with parameter $\alpha_{d}$, which is a vector with asymmetric topic priors for that document. Note that in this context each \textit{document} corresponds to individual sentences instead of full texts.
Then for each word $n$ in document $d$ a topic value is drawn: $z_{d,n}\sim Multi(\theta_{d})$, $z\in\{1 .. T\}$. Then a aspect-term/opinion switch variable is sampled: $y_{d,n}\sim Bernoulli(\pi_{d,n})$, $y\in \{A,O\}$. Depending on $y_{d,n}$, the word $w_{d,n}$ is emitted from the topic aspect terms distribution ($\phi_{z_{d,n},A}$) or else, a polarity value $v_{d,n}$ is sampled from $\Omega_{d}$ to choose if the word has to be drawn from $\phi_{z_{d,n},P}$ or $\phi_{z_{d,n},N}$ (positive and negative words respectively).

The model guides the topic and polarity modelling towards the desired values by biasing the hyper-parameters that govern the Dirichlet distributions from which the topics and words are sampled. In a standard LDA setting those hyper-parameters (commonly named $\alpha$ and $\beta$) are symmetric because no apriori information about the topic and word distributions is assumed.
In our model, these hyper-parameters are biased using a similarity calculation among the words of the domain corpus and the topic seed words of the initial configuration. This similarity measure is based on the cosine distance between the dense vector representation of the topic defining seeds and each word of the vocabulary. 
Such a dense vector representation of the words over a particular vocabulary, commonly referred as word embeddings, could be obtained using any distributional semantics approach, but in this work we stick to the well-known word2vec \citep{Mikolov2013}. Word embeddings are a very popular way of representing words as the input for a variety of machine learning techniques and are known for encoding interesting syntactic and semantic properties \citep{Mikolov2013lingRegu}. In this case, we exploit the semantic similarity among words that can be calculated using the cosine distance of the resulting word vectors. The similarity, $\similarity$, is the value between a word and a set of words (e.g. some topic defining seeds), and it is calculated using \ref{eq:similarity}.

\begin{equation}
\small
\similarity(w,t) = \argmax_{v\in t} \similarity(w,v)
\label{eq:similarity}
\end{equation}

Where $w$ is any word found in the domain corpus, $v$ is any of the seed words chosen to define topic $t$, and $\similarity$ stands for the cosine distance between two word vectors.

The $\alpha$ hyper-parameters control the topic probability distribution for each document as in the original LDA. But instead of having a single symmetric $\alpha$ value, each document has a biased $\alpha$ for each topic, based on semantic word similarity, as described in \ref{eq:alphas}.

\begin{equation}
\alpha_{d,t} = \frac{\sum\limits_{i}^{N_d}{\similarity(w_{d,i},t)}}
{\sum\limits_{t'}^{T}
	{\sum\limits_{i}^{N_d}{\similarity(w_{d,i},t')}}}*\alpha_{base}
\label{eq:alphas}
\end{equation}

On the other hand, the $\beta$ hyper-parameters, which control the distribution of words for each topic, are calculated in a similar way, as shown in \ref{eq:betas} and \ref{eq:betasPol}.

\begin{equation}
\beta_{t,w} = \similarity(w,t)*\beta_{base}
\label{eq:betas}
\end{equation}

\begin{equation}
\beta_{q,w} = \similarity(w,q)*\beta_{base} \quad q\in \{P,N\}
\label{eq:betasPol}
\end{equation}

Finally, the $\delta$ hyper-parameters control the polarity distribution for each document, and they are calculated for each document as shown in \ref{eq:deltas}.

\begin{equation}
\delta_{d,q} = \frac{\sum\limits_{i}^{N_d}{\similarity(w_{d,i},q)}}
{\sum\limits_{q'\epsilon\{P,N\}}
	{\sum\limits_{i}^{N_d}{\similarity(w_{d,i},q')}}}*\delta_{base}
\label{eq:deltas}
\end{equation}

In the formulas $w_{d,i}$ is the \textit{i-th} word of the document $d$, $N_d$ is the number of words in that document, $t$ is a topic from the set of defined topics $T$. Similarly $q$ is a pre-defined polarity words set, $P$ for positives and $N$ for negatives (in our experiments $P$ only contains \textit{excellent} and $N$ only contains \textit{horrible} for English, or their equivalents for other languages).

$\alpha_{base}$, $\beta_{base}$ and $\delta_{base}$ are configurable hyper-parameters, analogous to the symmetric $\alpha$ and $\beta$ in the original LDA model.

In addition to the bias of these hyper-parameters, the distribution $\pi$ that governs each binary aspect-term/opinion-word switching variable, $y$, is set from the pre-trained aspect-term/opinion-word classifier probabilities applied to each word and its context features as described in section \ref{sec:atow_separation}.

The posterior inference of the model is obtained via Gibbs sampling \citep{griffiths2004finding}.
Let $w_{d,n}$ be the $n$-th word of the $d$-th document, given the assignment of all other variables, its topic assignment $z_{d,n}$ is sampled using (\ref{equ:topic_sampling}).
Analogously, the aspect-term/opinion-word assignment $y_{d,n}$ and the polarity of the opinion-words, $v_{d,n}$ are sampled using (\ref{equ:atow_sampling}) and (\ref{equ:posneg_sampling}) respectively.

\begin{equation}
\small
p(z_{d,n}=t| z_{-d,n},y_{-d,n},v_{-d,n},\cdot)\propto \\
\frac{n^{t,A}_{w_{d,n}}+\beta^{t,A}_{w_{d,n}}}{\sum\limits_{v}^{V}n^{t,A}_{v}+\beta^{t,A}_{v}} \times
\frac{n^{t,P}_{w_{d,n}}+\beta^{t,P}_{w_{d,n}}}{\sum\limits_{v}^{V}n^{t,P}_{v}+\beta^{t,P}_{v}} \times \\
\frac{n^{t,N}_{w_{d,n}}+\beta^{t,N}_{w_{d,n}}}{\sum\limits_{v}^{V}n^{t,N}_{v}+\beta^{t,N}_{v}}
\times (n_{d,t}+\alpha_{d,t})
\label{equ:topic_sampling}
\end{equation}

\begin{equation}
\small
p(y_{d,n}=u|z_{d,n}=t,\cdot)\propto \\
\frac{n^{t,u}_{w_{d,n}}+\beta^{t,u}_{w_{d,n}}}{\sum\limits_{v}^{V}n^{t,u}_{v}+\beta^{t,u}_{v}}\times
\frac{exp(\lambda_{u}\times x_{d,n})}{\sum_{u'\in\{A,O\}}exp(\lambda_{u'}*x_{d,n})}
\label{equ:atow_sampling}
\end{equation}

\begin{equation}
\small
p(v_{d,n}=q|z_{d,n}=t,\cdot)\propto
\frac{n^{t,q}_{w_{d,n}}+\beta^{t,q}_{w_{d,n}}}{\sum\limits_{v'}^{V}n^{t,q}_{v'_{d,n}}+\beta^{t,q}_{v'_{d,n}}} \times (n_{d,q}+\delta_{d,q})
\label{equ:posneg_sampling}
\end{equation}

In these formulas, $n^{t,u}_{w_{d,n}}$ is the number of times the vocabulary term corresponding to $w_{d,n}$ has been assigned to topic $t$ and word-type $u\in \{A,O\}$ (i.e. Aspect-terms or Opinion-words), $n_{d,t}$ is the number of words in the document $d$ assigned to topic $t$, $\lambda_{u}$ are the pre-trained aspect-term/opinion-word classifier model weights for word-type $u$ and $x_{d,n}$ is the feature vector for $w_{d,n}$, composed by the Brown clusters of the context words. Analogously, $n^{t,q}_{w_{d,n}}$ is the number of times $w_{d,n}$ has been assigned to topic $t$ and polarity $q\in \{P,N\}$ and $n_{d,q}$ is the number of words in the document $d$ assigned to polarity $q$.

\section{Evaluation}
\label{sec:results}



We evaluate W2VLDA for the three different subtasks that it performs: topic (aspect) classification, sentiment classification, and aspect-term/opinion-word separation. First, we compare W2VLDA with other LDA-based methods. Then, we also evaluate W2VLDA in a multilingual ABSA dataset comparing its performance classifying topics (aspects) and sentiment with some supervised machine learning approaches trained on labelled data.

We show results for several datasets, demonstrating how the system works for different languages and domains just by changing the topic configuration, composed of a single seed word per each desired topic, language and domain.


\begin{table}[!t]
\scriptsize
  \centering
   \resizebox{\linewidth}{!}{%
    \begin{tabular}{|p{2.4cm}|p{1.5cm}|p{2.8cm}|p{2.8cm}|p{2.8cm}|}
    \hline
    Language:Domain & \multicolumn{1}{c|}{Domain topic} & \multicolumn{1}{c|}{Aspect-terms} & \multicolumn{1}{c|}{Positive words} & \multicolumn{1}{c|}{Negative words} \bigstrut\\
    \hline
    \hline
    \multirow{3}[6]{*}{\parbox{2.4cm}{English:\\ restaurant reviews}} &\vspace{.02\baselineskip} Food  & \vspace{.02\baselineskip} chicken, beef, pork, tuna, egg, onions, shrimp, curry & \vspace{.02\baselineskip} moist, goat, smoked, seared, roasted, red, crispy, tender & \vspace{.02\baselineskip} undercooked, dry, drenched, overcooked, soggy, chewy \bigstrut\\
\cline{2-5}          & \vspace{.02\baselineskip} Service & \vspace{.02\baselineskip} staff, workers, employees, chefs, hostess, manager, owner & \vspace{.02\baselineskip} helpful, polite, knowledgeable, efficient, prompt, attentive & \vspace{.02\baselineskip} inattentive, rude, unfriendly, wearing, making, packed \bigstrut\\
\cline{2-5}          & \vspace{.02\baselineskip} Ambiance & \vspace{.02\baselineskip} lighting, wall, interior, vibe, concept, ceilings, setting, decor & \vspace{.02\baselineskip} modern, beautiful, chic, nice, trendy, cozy, elegant, cool & \vspace{.02\baselineskip} bad, loud, uninspired, expensive, big, noisy, dark, cramped \bigstrut\\
    \hline
    \hline
    \multirow{3}[6]{*}{\parbox{2.4cm}{English:\\ electronic devices reviews}} &\vspace{.02\baselineskip} Warranty &\vspace{.02\baselineskip} warranty, support, repair, service, answer, center, policy &\vspace{.02\baselineskip} worked, lucky, owned, big, exchange, extended, longer &\vspace{.02\baselineskip} called, contact, broken, faulty, defective, expired, worthless \bigstrut\\
\cline{2-5}          &\vspace{.02\baselineskip} Design &\vspace{.02\baselineskip} plastic, wheel, style, handle, pocket, design, exterior, wheels &\vspace{.02\baselineskip} adjustable, clean, good, versatile, attractive, lightweight, stylish &\vspace{.02\baselineskip} ugly, odd, awkward, tight, felt, weird, cute, stupid, flimsy \bigstrut\\
\cline{2-5}          &\vspace{.02\baselineskip} Price &\vspace{.02\baselineskip} money, store, item, bucks, price, regret, deal, gift &\vspace{.02\baselineskip} paying, reasonable, penny, worth, delivered, stars, inexpensive,  & \vspace{.02\baselineskip}disappointed, paid, cheaper, skeptical, pricey, overpriced \bigstrut\\
    \hline
    \end{tabular}%
    }
    \caption{Resulting topic words distributions for English in two different domains. The topics are automatically split into three different word distributions: topic aspect terms, topic positive words and topic negative words.}
  \label{tab:topic_words_example_en}%
\end{table}%

\begin{table}[!t]
\scriptsize
  \centering
  \resizebox{\linewidth}{!}{%
    \begin{tabular}{|p{2.4cm}|p{1.5cm}|p{2.8cm}|p{2.8cm}|p{2.8cm}|}
    \hline
    Language:Domain & \multicolumn{1}{c|}{Domain topic} & \multicolumn{1}{c|}{Aspect-terms} & \multicolumn{1}{c|}{Positive words} & \multicolumn{1}{c|}{Negative words} \bigstrut\\
    \hline
    \hline
    \multirow{3}[6]{*}{\parbox{2.4cm}{Spanish:\\ restaurant reviews}} & \vspace{.02\baselineskip} Food  & \vspace{.02\baselineskip} crema, tartar, ensaladas, sopa, brasa, patatas, salsas, alcachofas & \vspace{.02\baselineskip} caprese, sublime, destacar, casera, tierna, trufada, ahumada & \vspace{.02\baselineskip} aguada, mojar, congeladas, quemadas, fritos, rancias, reseco \bigstrut\\
\cline{2-5}          &\vspace{.02\baselineskip} Service &\vspace{.02\baselineskip} camareros, camarero, maitre, due\~no, encargado, metre &\vspace{.02\baselineskip} eficiente, eficaz, atentos, correcta, cercano, diligente & \vspace{.02\baselineskip}lento, p\'esimo, desagradable, prepotente, maleducado \bigstrut\\
\cline{2-5}          & \vspace{.02\baselineskip}Ambiance & \vspace{.02\baselineskip} toques, atm\'osfera, material, mobiliario, bancos, modernidad &  \vspace{.02\baselineskip}tranquilo, relajado, c\'alido,  buena, amplio, luminoso, precioso, & \vspace{.02\baselineskip} cutre, insoportable, peque\~no, tanta, oscuro, poca, normalita \bigstrut\\
    \hline
    \hline
    \multirow{3}[6]{*}{\parbox{2.4cm}{French:\\ hotel reviews}} &\vspace{.02\baselineskip} Food  &\vspace{.02\baselineskip} nourriture, sauce, produits, p\^ate, bouffe, saveur, risotto & \vspace{.02\baselineskip} raisonnable, michelin,  excellents, merveilleuse, veritable, superbe & \vspace{.02\baselineskip} correcte,  cuit, idem, passable, excessif, moleculaire, difficile \bigstrut\\
\cline{2-5}          & \vspace{.02\baselineskip}Staff & \vspace{.02\baselineskip} personnel, \`ecoute, staff, gentillesse, concierge, membres & \vspace{.02\baselineskip} sympathique, attentionn\'e, efficace, comp\`etent, professionnel & \vspace{.02\baselineskip} d\`eplorable,  antipathique, d\`ebord\`e, distant, constamment \bigstrut\\
\cline{2-5}          & \vspace{.02\baselineskip} Ambiance & \vspace{.02\baselineskip} impression, couloirs, odeurs, personnages, hiver, escaliers & \vspace{.02\baselineskip} vieillissant, grand, r\`enov\`e, boone, typiquement, cosy, agr\`eablement & \vspace{.02\baselineskip} froide, v\`etuste, forte, incendie, bruyants, inexistante, compl\`ete \bigstrut\\
    \hline
    \end{tabular}%
    }
    \caption{Resulting topic words distributions for two Spanish and French and for different domains. The topics are automatically split into three different word distributions: topic aspect terms, topic positive words and topic negative words.}
  \label{tab:topic_words_example_es_fr}%
\end{table}%

For instance, table \ref{tab:topic_words_example_en} shows some of the resulting words for several domains (restaurants and electronic devices), topics (food, service, ambience for restaurants, and warranty, design and price for electronic devices) for English customer reviews, including the automatic separation of aspect-terms from positive and negative words per topic.
Table \ref{tab:topic_words_example_es_fr} shows the equivalent information for restaurants and hotel reviews in Spanish and French.

\begin{table}[!t]
\scriptsize
  \centering
  \resizebox{\linewidth}{!}{%
    \begin{tabular}{|p{1.2cm}|p{1.2cm}|p{10cm}|}
    \hline
    \multicolumn{1}{|p{1.2cm}|}{Lang: Domain} & \multicolumn{1}{p{1.2cm}|}{Domain Topic} & Examples of sentences with high posterior probability for different topics\bigstrut\\
    \hline
    \hline
    \multicolumn{1}{|p{1.2cm}|}{\multirow{3}[6]{*}{\parbox{1.2cm}{English:\\ restaurant reviews}}} & Food  & {\parbox{10cm}{\vspace{.5\baselineskip}Appetizer was grilled pizza dough topped with fig jam, prosciutto, arugula, cherry tomatoes... \vspace{.5\baselineskip}
    		
Four of us enjoyed sizziling rice seafood soup, the most savory garlic string beans.\vspace{.5\baselineskip}}} \bigstrut\\
\cline{2-3}          & Service & {\parbox{10cm}{\vspace{.5\baselineskip}Seated promptly, waiter arrived at 6:10 brought us our drink order 6:15.\vspace{.5\baselineskip}
		
Bartenders are friendly and quick to be helpful \vspace{.5\baselineskip}}}\bigstrut\\
\cline{2-3}          & Ambiance & {\parbox{10cm}{\vspace{.5\baselineskip}The atmosphere as a restaurant though is very nice: cute decor, quieter, and dim lighting...\vspace{.5\baselineskip}
		
The ambiance of the restaurant is very nice, the decor and lighting set a great atmosphere\vspace{.5\baselineskip}}} \bigstrut\\
    \hline
    \hline
    \multicolumn{1}{|p{1.2cm}|}{\multirow{3}[6]{*}{\parbox{1.2cm}{Spanish:\\ restaurant reviews}}} & Food  & {\parbox{10cm}{\vspace{.5\baselineskip}Probamos las croquetas melosas de jam\'on, milhoja de tomate y mozzarella con salsa de miel.\vspace{.5\baselineskip}
    		
Pat\'e de perdiz, tartar de bonito, steak tartar, pat\'e de cabracho, brocheta de pollo y postres\vspace{.5\baselineskip}}} \bigstrut\\
\cline{2-3}          & Service & {\parbox{10cm}{\vspace{.5\baselineskip}El servicio a los clientes deja bastante que desear\vspace{.5\baselineskip}
		
El trato es magn\'ifico, camareros muy simp\'aticos y amables, un trato educado y exquisito\vspace{.5\baselineskip}}} \bigstrut\\
\cline{2-3}          & Ambiance & {\parbox{10cm}{\vspace{.5\baselineskip}Cena agradable en un lugar de ambiente tranquilo, cosmopolita, con buena m\'usica\vspace{.5\baselineskip}
		
 El local es feo decorado como un bar de carretera en EEUU o un autob\'us\vspace{.5\baselineskip}}} \bigstrut\\
    \hline
    \hline
    \multirow{3}[6]{*}{\parbox{1.2cm}{French:\\ hotel reviews}} & Staff & {\parbox{10cm}{\vspace{.5\baselineskip}Service de qualit\`e et personnel extr\^emement agreable, aux petits soins, disponible et serviable!\vspace{.5\baselineskip}
    		
Le personnel est r\`eactif, serviable, disponible, toujours pr\^et \`a r\`epondre aux attentes des clients.\vspace{.5\baselineskip}}} \bigstrut\\
\cline{2-3}          & Ambiance & {\parbox{10cm}{\vspace{.5\baselineskip}L'hotel est une attraction en soi, il y a un adventure park a l'interieur, on se croirait a disneyland.\vspace{.5\baselineskip}
		
Le b\^atiment a un certain charme, certaines tapisseries sont d\`efra\^ichies, se sent londonien\vspace{.5\baselineskip}}} \bigstrut\\
\cline{2-3}          & Location & {\parbox{10cm}{\vspace{.5\baselineskip}A 5 minutes \`a pied de buckingham palace et saint james park , 10 \`a 15 minutes de big ben.\vspace{.5\baselineskip}
		
Hotel \`a 15 min de la gare \`a pied, à 15 min d'oxford street, \`a 40 min du centre ville \`a pied.\vspace{.5\baselineskip}}} \bigstrut\\
    \hline
    \end{tabular}%
    }
      \caption{Some examples of sentences with the highest posterior probability for several topics, domains and languages.}
  \label{tab:topic_sentences_example}%
\end{table}%

Likewise, table \ref{tab:topic_sentences_example} shows examples of sentences classified under different topics (food, service, ambience for restaurants, and staff, ambience and location for hotels) for several domains (restaurants and hotels) and languages (Spanish and French).

\subsection{Resources and experimental setting}
\label{sec:resources}

In order to evaluate W2VLDA, we use the following resources. For topic classification we use the dataset from \citep{Ganu2009} which contains restaurant reviews labelled with domain-related categories (e.g. “food”, “staff”, “ambience”) for English.
For sentiment classification, we use the Laptops and DIGITAL-SLR dataset \citep{Jo2011}, consisting of English reviews of electronic products with their corresponding 5-star rating.

Additional multilingual experiments have been performed using the SemEval-2016 task 5 datasets \citep{pontiki-EtAl:2016:SemEval}. In particular, the restaurant reviews datasets which are labelled with domain-related categories and polarity for six languages.

In order to compute the topic model and the word embeddings, we have automatically gathered additional customer reviews about restaurants from some popular customer review websites. These unlabelled domain corpora consist of a few thousand restaurant reviews in English, Spanish, French and Dutch.

We use word2vec to compute the word embeddings that are used for the word similarity calculation. In particular, we use the Apache Spark MLlib \footnote{\url{http://spark.apache.org/mllib/}} implementation with default parameters to compute the domain-based word embeddings.

Table \ref{tab:seeds_restaurants} shows the topic definition used in the experiments for the domain of restaurants, just one word per topic. Unless stated otherwise, the polarity seeds for every domain are \textit{excellent} and \textit{horrible} or their equivalents in other languages.

The values for $\alpha_{base}$, $\beta_{base}$ and $\delta_{base}$ mentioned in \ref{sec:topic_model}, which play a similar role to $\alpha$ and $\beta$ in the original LDA, are set to the values commonly recommended in the literature \citep{griffiths2004finding}: 50/T for $\alpha_{base}$ and $\delta_{base}$ being T the number of topics, and 0.01 for $\beta_{base}$. The topic modelling process runs for 500 iterations in every experiment with a burn-in period of 100 iterations and a sampling lag of 10 iterations.

\subsection{Comparison with other LDA based approaches}
\label{sec:topic_eval}

First, we evaluate W2VLDA in a topic classification setting using the restaurant reviews dataset from \citep{Ganu2009}.  This dataset contains few thousand reviews from restaurants, classified into several categories but the authors report results only for the three main categories: \textit{food}, \textit{ambience} and \textit{staff}. We compare W2VLDA against the results reported in \citep{Zhao2010} for two LDA-based approaches, LocLDA \citep{Brody2010} and ME-LDA \citep{Zhao2010}.


LocLDA and ME-LDA are LDA-based approaches, and thus, unsupervised. But the results reported in the experiment involved some supervision as described in \cite{Zhao2010}. First, the authors computed a topic model of 14 topics. Then the authors examine each topic and manually set a label according to their judgment. W2VLDA provides already named topics at the end of the process, so no manual topic inspection and labelling are required.
In order to assign a topic label to a particular sentence, we use the resulting topic distribution for that sentence ($\theta_{d}$) selecting the topic with highest posterior probability.


\begin{table}[t]
	
	\centering
	\resizebox{\linewidth}{!}{%
		\begin{tabular}{|l|c|c|c|c|c|c|c|c|c|c|c|c|}
			
			\hline
			\multicolumn{1}{|c|}{\multirow{3}[5]{*}{Method}} & \multicolumn{12}{c|}{Topics} \bigstrut\\
			\cline{2-13}          & \multicolumn{3}{c|}{Staff} & \multicolumn{3}{c|}{Food} & \multicolumn{3}{c|}{Ambiance} & \multicolumn{3}{c|}{Overall} \bigstrut\\
			\cline{2-13}          & Prec. & Rec.  & F-1   & Prec. & Rec.  & F-1   & Prec. & Rec.  & F1    & Prec. & Rec.  & F1 \bigstrut[t]\\ \hline
			LocLDA & \textbf{0.80} & 0.59  & 0.68  & 0.90  & 0.65  & 0.75  & 0.60  & 0.68  & 0.64  & 0.77  & 0.64  & 0.69 \bigstrut[b]\\
			\hline
			ME-LDA & 0.78  & 0.54  & 0.64  & 0.87  & \textbf{0.79} & \textbf{0.83} & \textbf{0.77} & 0.56  & \textbf{0.65} & \textbf{0.81} & 0.63  & 0.70 \bigstrut\\
			\hline
			W2VLDA & 0.61  & \textbf{0.86} & \textbf{0.71} & \textbf{0.96} & 0.69  & 0.81  & 0.55  & \textbf{0.75} & 0.63  & 0.70  & \textbf{0.77} & \textbf{0.72} \bigstrut\\
			\hline
			
		\end{tabular}%
	}
	\caption{Comparison against other LDA based approaches on restaurants domain}
	\label{tab:zhao-comparison}%
	
\end{table}%

Table \ref{tab:zhao-comparison} shows the results of the experiment and the comparison with the other systems. Despite not requiring human intervention to relabel the obtained topics unlike the other two systems, W2VLDA obtains slightly better overall results.

We also evaluate the ability of W2VLDA to assign correct polarities to customer reviews. We use the estimated polarity distribution of a sentence ($\Omega_{d}$) to assign to a review the polarity with the highest probability. We compare our polarity classification results with respect to those from JST \citep{Lin2012}, ASUM \citep{Jo2011} and HASM \citep{KimSuin2013}. The evaluation runs over the laptops and digital SLRs subset obtained from the Amazon Electronics dataset\footnote{Available at \url{http://uilab.kaist.ac.kr/research/WSDM11/}}. As explained at \citep{KimSuin2013} two datasets are used, a \textit{small} dataset containing 1000 reviews with 1 star rating (strong negative) and 1000 5 stars (strong positive), and a \textit{large} dataset with additional 1000 reviews of 2 stars (negative) as well as 1000 reviews of 4 stars (positive).
The baseline consists of a simple polarity seed word count, using the polarity seed words from \citep{turney2003measuring}, assigning to the sentence the polarity with the greatest proportion. As stated in previous sections, W2VLDA uses just a single polarity seed for each sentiment polarity, \textit{excellent} and \textit{horrible} respectively.

\begin{figure*}[t]
	
	\centering
	\includegraphics[width=0.8\textwidth]{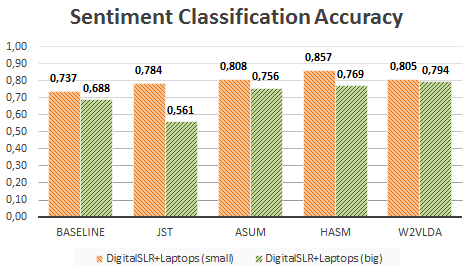}
	\caption{Sentiment classification accuracy comparison with other LDA based approaches in a electronic devices reviews dataset}
	\label{figure:sentiment_eletronics}
	
\end{figure*}

Figure \ref{figure:sentiment_eletronics} shows the result of this comparison. W2VLDA obtains comparable results for the small dataset and better results for the big dataset despite using only a single seed word to define each polarity.


\subsection{Multilingual evaluation on SemEval2016 dataset}

We use the SemEval 2016 task 5 datasets \citep{pontiki-EtAl:2016:SemEval} in order to perform a multilingual evaluation of W2VLDA. SemEval 2016 datasets consist of restaurant reviews in several languages. The reviews are split by sentence and labelled with the explicit aspect term mentions, the coarse-grained category they belong to, and the polarity for that category.

\begin{figure*}[t]
	
	\centering
	\includegraphics[width=1.0\textwidth]{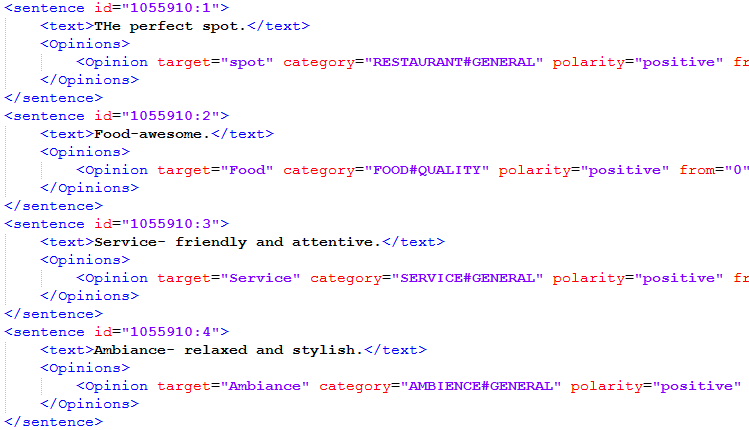}
	\caption{SemEval 2016 task 5 restaurants dataset example (for English).}
	\label{figure:semeval_dataset_example}
	
\end{figure*}

SemEval 2016 restaurants datasets are annotated for six coarse-grained categories: food, service, ambience, drinks, location, and restaurant. The last category, \textit{restaurant} acts as a miscellaneous category that is used when the sentence does not refer to any other specific category but to the restaurant as a whole. Such an abstract concept cannot be represented by a seed word, so we omit this category from the evaluation.
To avoid ambiguities and simplify the classification of a sentence, we only keep sentences with a single category label. Finally, since the categories \textit{drinks} and \textit{location} have very little representation in the datasets (below the 5\% of the instances), we keep only the three main categories: \textit{food}, \textit{service} and \textit{ambience}.

Table \ref{tab:topic_distribution_semeval2016_filtered} and table \ref{tab:polarity_distribution_semeval2016_filtered} show the distribution of categories and polarities respectively for the resulting datasets, for four languages: English, Spanish, French and Dutch.

\begin{table}[t]
	\centering
	\footnotesize
	\begin{tabular}{|l|r|r|r|r|}
		\hline
		& \multicolumn{1}{l|}{\textbf{EN}} & \multicolumn{1}{l|}{\textbf{ES}} & \multicolumn{1}{l|}{\textbf{FR}} & \multicolumn{1}{l|}{\textbf{NL}} \bigstrut\\
		\hline
		Food  & 486   & 364   & 370   & 374 \bigstrut\\
		\hline
		Service & 328   & 233   & 290   & 350 \bigstrut\\
		\hline
		Ambience & 110   & 145   & 98    & 117 \bigstrut\\
		\hline
		Total & 924   & 742   & 758   & 841 \bigstrut\\
		\hline
	\end{tabular}%
	\caption{SemEval 2016 dataset category distribution after filtering unwanted categories and sentences with more than one annotation.}
	\label{tab:topic_distribution_semeval2016_filtered}%
\end{table}%

\begin{table}[t]
	\centering
	\footnotesize
	\begin{tabular}{|l|r|r|r|r|}
		\hline
		& \multicolumn{1}{l|}{\textbf{EN}} & \multicolumn{1}{l|}{\textbf{ES}} & \multicolumn{1}{l|}{\textbf{FR}} & \multicolumn{1}{l|}{\textbf{NL}} \bigstrut\\
		\hline
		Positive & 551   & 417   & 300   & 405 \bigstrut\\
		\hline
		Negative & 326   & 273   & 413   & 369 \bigstrut\\
		\hline
		Total & 877   & 690   & 713   & 774 \bigstrut\\
		\hline
	\end{tabular}%
	\caption{SemEval 2016 dataset polarity distribution after filtering unwanted categories and sentences with more than one annotation.}
	\label{tab:polarity_distribution_semeval2016_filtered}%
\end{table}%

Since W2VLDA is a topic modelling, it needs a reasonable amount of domain documents to build the statistical model. To cope with this requirement, we have implemented a script to automatically extract restaurant reviews of the required languages from an online customer reviews website. Due to copyright permissions, we cannot share these reviews, but table \ref{tab:downloaded_reviews} shows the number of reviews used to feed the algorithm. The polarity mentioned on the table is based on the number of the stars from the 5-star rating (as usual, 1-2 stars meaning negative and 4-5 starts meaning positive).
As it can be observed in the table, for some languages the script has not found an equal number of positive and negative reviews. We tried to compensate this fact with oversampling, to pair the number of positive and negative reviews before running the algorithm. In this case we oversample negative examples for each language until they equal in number the positive ones (i.e. 10k). Note that in the case of Dutch this may lead to an excessive oversampling due to the small number of available negatives examples. Also note that these polarities are just to get an insight of the polarity distribution of the datasets, but they are not used for any sort of supervised training.

\begin{table}[t]
	\footnotesize
	\centering
	\begin{tabular}{|l|r|r|r|r|}
		\hline
		\multicolumn{5}{|c|}{Restaurant customer reviews downloaded from a website} \bigstrut\\
		\hline
		& \multicolumn{1}{l|}{EN} & \multicolumn{1}{l|}{ES} & \multicolumn{1}{l|}{FR} & \multicolumn{1}{l|}{NL} \bigstrut\\
		\hline
		Positives (4 or 5 stars) & 10000 & 10000 & 10000 & 10000 \bigstrut\\
		\hline
		Negatives (1 or 2 stars) & 10000 & 8400  & 5500  & 830 \bigstrut\\
		\hline
		Total reviews & 20000 & 18400 & 15500 & 10830 \bigstrut\\
		\hline
	\end{tabular}%
	\caption{Downloaded reviews distribution per language and polarity (using 5-star rate). The automatic script could not find the same number of negative reviews for all the languages. We try to alleviate this problem oversampling negatives reviews.}
	\label{tab:downloaded_reviews}%
\end{table}%

The evaluation experiment is done as follows. For each language, we use the downloaded reviews to run the algorithm. It includes calculating the domain word embeddings, Brown clusters and the topic model estimation.
Using the generated model for each language the topic and polarity distributions, $\theta$ and $\Omega$, are estimated for each of the sentences of the evaluation set.
The topic with the highest probability in the estimated topic distribution for that sentence is assigned as the category label (i.e. domain aspect).
Analogously, the polarity with the highest probability in the estimated polarity distribution for that sentence is assigned as the polarity label.
The assigned category is compared to the gold category, and the accuracy (ratio of correctly labelled examples) is calculated. The same process is followed to calculate the polarity classification accuracy.

The obtained accuracy is compared to several baselines.
First, two supervised baselines are used.
One is a Naive-Bayes classifier (NB), trained using the labelled sentences. The sentences are transformed to bag-of-words vectors with a vocabulary size of 80k words and normalised using tf-idf weights.
The other supervised baseline is a Multilayer Perceptron algorithm (MLP), with two hidden layers, and the same tf-idf vector as input.
Another baseline is the majority baseline, that shows the accuracy that can be obtained in the case of choosing the most frequent class. This is only to ensure that the datasets are not excessively unbalanced and the algorithms are really learning relevant information.
Finally, the last baseline (W2VLDA\_NO) is the same W2VLDA but replacing the word-embeddings similarity mechanism to bias the topic modelling hyper-priors. Instead of using the word-embedding similarity to calculate a bias for every word, only the configured seed words receive a strong bias for their corresponding topic or polarity.


\begin{table}[t]
	\centering
	\footnotesize
	\begin{tabular}{|l|c|c|c|c|}
		\hline
		\multicolumn{5}{|c|}{\textbf{Domain aspect classification}} \bigstrut\\
		\hline
		& \textbf{EN} & \textbf{ES} & \textbf{FR} & \textbf{NL} \bigstrut\\
		\hline
		NB    & 0.492 & 0.497 & 0.472 & 0.457 \bigstrut\\
		\hline
		MLP   & 0.554 & 0.564 & 0.496 & 0.464 \bigstrut\\
		\hline
		Majority baseline & 0.333 & 0.333 & 0.333 & 0.333 \bigstrut\\
		\hline
		W2VLDA\_NO & 0.313 & 0.374 & 0.356 & 0.315 \bigstrut\\
		\hline
		W2VLDA & \textbf{0.781} & \textbf{0.633} & \textbf{0.586} & \textbf{0.473} \bigstrut\\
		\hline
	\end{tabular}%
	\caption{Domain aspect classification results. NB and MLP are the supervised baselines, NaiveBayes and MultiLayer perceptron respectively. Majority baseline shows which would be the result of simply choosing the most frequent class. W2VLDA\_NO is the proposed approach without word embeddings. W2VLDA is the proposed approach.
    }
	\label{tab:results_topic_semeval}%
\end{table}%

Table \ref{tab:results_topic_semeval} shows the evaluation results for the domain aspects classification (food, service, ambience).
Since the evaluation datasets are not completely balanced for each of the domain aspects (see table \ref{tab:topic_distribution_semeval2016_filtered}), we run the evaluation on several balanced subsets created by random sampling the base datasets for each language. Each balanced subset contains 100 sentences from each domain aspect. We do this five times generating five different subsets, and we use these subsets to evaluate the baselines and W2VLDA.
The results on each individual subset are obtained using the average accuracy applying a 10-fold cross validation.
We calculate the average and standard deviation of the results on each subset to perform a t-test of statistical significance. W2VLDA outperforms the baselines with a 95\% of confidence for all the languages except for Dutch, which despite obtaining better results than the baselines it only achieves a 80\% on confidence in the statistical significance test.



\begin{table}[t]
	\centering
	\footnotesize
	\begin{tabular}{|l|c|c|c|c|}
		\hline
		\multicolumn{5}{|c|}{\textbf{Sentiment polarity classification}} \bigstrut\\
		\hline
		& \textbf{EN} & \textbf{ES} & \textbf{FR} & \textbf{NL} \bigstrut\\
		\hline
		NB    & 0.672 & 0.577 & 0.587 & 0.563 \bigstrut\\
		\hline
		MLP   & 0.711 & 0.602 & 0.583 & 0.577 \bigstrut\\
		\hline
		Majority baseline & 0.500 & 0.500 & 0.500 & 0.500 \bigstrut\\
		\hline
		W2VLDA\_NO & 0.531 & 0.552 & 0.534 & 0.523 \bigstrut\\
		\hline
		W2VLDA & \textbf{0.773} & \textbf{0.723} & \textbf{0.628} & \textbf{0.623} \bigstrut\\
		\hline
	\end{tabular}%
		\caption{Sentiment polarity classification results. NB and MLP are the supervised baselines, NaiveBayes and MultiLayer perceptron respectively. Majority baseline shows which would be the result of simply choosing the most frequent class. W2VLDA\_NO is the proposed approach without word embeddings. W2VLDA is the proposed approach.}
	\label{tab:results_polarity_semeval}%
\end{table}%

Table \ref{tab:results_polarity_semeval} shows the evaluation results for the polarity classification (positive and negative).
The calculation of the results and the statistical significance tests have been performed in the same way than for the domain aspect classification. Again, W2VLDA outperforms the baselines with a 95\% on confidence in the statistical test, except for Dutch. A possible reason for this is that the oversampling performed for the unlabelled Dutch reviews for the topic modelling was excessive, or the data contained in it was less representative than for other languages (see table \ref{tab:downloaded_reviews}).
Studying which are the lower bounds of the required amount of data would be an interesting problem that we let for future research.


\subsection{Assessing the seed words impact}
\label{sec:seed_words_impact}

Since the proposed approach heavily relies on the seed words (i.e. seeds words are the only source of supervision to guide the algorithm to the desired goal), it is interesting to evaluate the impact of different seed words and their combination.

We perform some experiments for English using the SemEval 2016 restaurant reviews dataset and several combinations of seed words for the target domain aspects and sentiment polarities.
In the first experiment group, for each run we only change the seed words that define the domain aspects. The polarity seeds remain the same. 

\begin{table}[t]
	\centering
	\footnotesize
	
	\resizebox{\linewidth}{!}{%
		\begin{tabular}{|l|r|r|}
			\hline
			\textbf{Aspects:\{FOOD\},\{SERVICE\},\{AMBIENCE\}} & \multicolumn{1}{l|}{\textbf{Aspects acc.}} & \multicolumn{1}{l|}{\textbf{Polarity acc}} \bigstrut\\
			\hline
			\hline
			\{food\},\{service\},\{ambience\} & 0,709 & 0,738 \bigstrut\\
			\hline
			\{chicken\},\{staff\},\{atmosphere\} & 0,653 & 0,729 \bigstrut\\
			\hline
			\{burger\},\{waiter\},\{d\'ecor\} & 0,662 & 0,731 \bigstrut\\
			\hline
			\{food,chicken\},\{service,staff\},\{ambience,atmosphere\} & 0,735 & \textbf{0,742} \bigstrut\\
			\hline
			\{food,burger\},\{service,waiter\},\{ambience,d\'ecor\} & 0,724 & 0,721 \bigstrut\\
			\hline
			\{chicken,burger\},\{staff,waiter\},\{atmosphere,d\'ecor\} & 0,673 & 0,725 \bigstrut\\
			\hline
			All the 3 seeds for every aspect & \textbf{0,761} & 0,722 \bigstrut\\
			\hline
			\hline
			Average   & 0,702 & 0,730 \bigstrut\\
			\hline
			Standard deviation   & 0,041 & 0,008 \bigstrut\\
			\hline
		\end{tabular}%
	}
	\caption{Impact of different seed words combination for the domain aspect classification.}
	\label{tab:seeds_aspect_impact}%
\end{table}%

We use three different seed words for each domain aspect, in particular: \textit{food}, \textit{chicken} and \textit{burger} for domain aspect \textit{FOOD}; \textit{service}, \textit{staff} and \textit{waiter} for domain aspect \textit{SERVICE}; and \textit{ambience}, \textit{atmosphere} and \textit{d\'ecor} for domain aspect \textit{AMBIENCE}. We try different permutations and combinations of the seed words, including pairs of seed words for each domain aspect, and finally also the combination of the three seed words together.
Table \ref{tab:seeds_aspect_impact} show the results for this experiment. The results show that the accuracy is stable across all the combinations regardless of the chosen seed words. As expected, some combinations perform better than others but overall the average if high and the standard deviation is below 5\% of the accuracy. The best result is obtained using all the seed words at the same time. This last fact is not surprising, since with more seeds the semantic coverage to guide the algorithm to the desired domain aspects is increased (as long as the seed words are semantically coherent with the domain aspect they are defining).

Another fact that can be observed in the table is that domain aspect seed words do not affect the polarity results, as it would be expected. The polarity results show minor variations among the experiments, but the standard deviation is only a 0.8\% of the accuracy.

\begin{table}[t]
	\centering
	\footnotesize
	\resizebox{\linewidth}{!}{%
		\begin{tabular}{|l|r|r|}
			\hline
			\textbf{Polarity:\{POSITIVE\},\{NEGATIVE\}} & \multicolumn{1}{l|}{\textbf{Aspects acc.}} & \multicolumn{1}{l|}{\textbf{Polarity acc.}} \bigstrut\\
			\hline
			\hline
			\{excellent\},\{horrible\} & 0,701 & 0,724 \bigstrut\\
			\hline
			\{terrific\},\{terrible\} & 0,712 & 0,736 \bigstrut\\
			\hline
			\{awesome\},\{awful\} & 0,691 & 0,745 \bigstrut\\
			\hline
			\{nice\},\{poor\} & 0,704 & 0,735 \bigstrut\\
			\hline
			\{good\},\{bad\} & 0,684 & 0,712 \bigstrut\\
			\hline
			\{affordable\},\{expensive\} & \textbf{0,716} & 0,729 \bigstrut\\
			\hline
			\{excellent,terrific\},\{horrible,terrible\} & 0,683 & 0,726 \bigstrut\\
			\hline
			\{excellent,terrific,awesome\},\{horrible,terrible,awful\} & 0,692 & \textbf{0,747} \bigstrut\\
			\hline
			\hline
			Average   & 0,698 & 0,732 \bigstrut\\
			\hline
			Standard deviation   & 0,012 & 0,012 \bigstrut\\
			\hline
		\end{tabular}%
	}
	\caption{Impact of different polarity seeds words for the sentiment polarity classification.}
	\label{tab:seeds_polarity_impact}%
\end{table}%

Analogously to the domain aspect seed words, we have performed some experiments with the polarity words. We have tested several combinations of seeds with opposed polarity: excellent - horrible, awesome - awful, etc. Table \ref{tab:seeds_polarity_impact} show the results. Even with seed words of less extreme polarity, like good - bad, the results are quite stable. We also test combining more than a single word for each polarity, and as the results table shows, combining the three seed words for each polarity achieves the best result. The standard deviation for all the experiment runs is just a 1.2\% of the accuracy. Similarly to what was observed for the domain aspects, the polarity seed words do not seem to affect the domain aspect classification accuracy, with only a 1.2\% on standard deviation for all the runs.

Finally, in order to perform a sanity check to evaluate if the sentiment polarity classification is really depending on the correct selection of the polarity seed words, we perform two more runs using misleading words as polarity seeds. In particular, we use \textit{cat} and \textit{waitress} for positives and \textit{dog} and \textit{waiter} for negatives. The use of these words as polarity seeds is obviously incorrect, and what we want to check is if using such meaningless words (for polarity) leads to bad polarity classification results. Table \ref{tab:seeds_nonsense_polarity_words} shows the results for this experiment, confirming that the election of representative polarity seed words is relevant to correctly guide the algorithm.

\begin{table}[t]
	\centering
	\footnotesize
	\resizebox{\linewidth}{!}{%
	\begin{tabular}{|l|r|r|}
		\hline
		\textbf{Polarity:\{POSITIVE\},\{NEGATIVE\}} & \multicolumn{1}{l|}{\textbf{Aspects acc.}} & \multicolumn{1}{l|}{\textbf{Polarity acc.}} \bigstrut\\
		\hline
		\hline
		\{cat\},\{dog\} & 0,642 & 0,447 \bigstrut\\
		\hline
		\{waitress\},\{waiter\} & 0,635 & 0,419 \bigstrut\\
		\hline
	\end{tabular}%
}
	\caption{Results using misleading words as polarity seeds to check to which extent the sentiment polarity classification depends on the validity of the chosen polarity seeds.}
	\label{tab:seeds_nonsense_polarity_words}%
\end{table}%


\subsection{Aspect-term/Opinion-word separation evaluation}
\label{sec:atow_separation_eval}
Finally we experiment with the aspect-term and opinion-word separation.
As described in section \ref{sec:atow_separation}, W2VLDA models the domain words into separated word distributions: aspect terms or opinion words. 

In order to evaluate the accuracy of this words separation, we use Bing Liu's polarity lexicon for English \citep{Hu2004}. Since sentiment lexicons contain words bearing some specific sentiment, we treat the words contained in this lexicon as a ground-truth for opinion-words. In addition, we use the gold aspect-terms labelled in the SemEval 2016 dataset as a ground-truth for aspect-terms.

The experiment now consists of running the W2VLDA again on the restaurant review dataset and counting how many times a word from the opinion words ground-truth is classified as an opinion word, and how many times each word from the aspect terms ground-truth is classified as an aspect term.
Then the proportion of correct assignments is calculated.
If the automatic aspect-term / opinion-word separation is correct, the proportion of opinion words and aspect terms correctly classified should be high.

\begin{figure*}[t]
	\centering
	\includegraphics[width=0.8\textwidth]{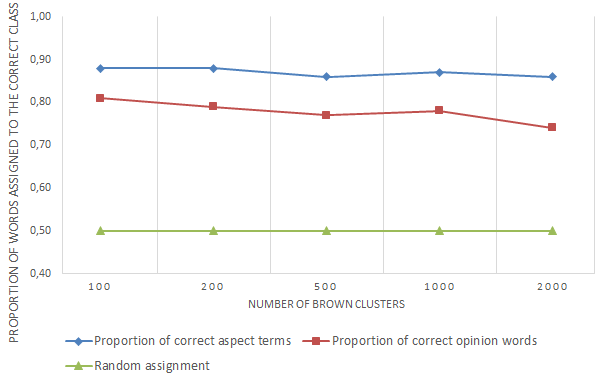}
	\caption{Result of aspect term and opinion word separation for English. Each point indicates the correct proportion (percentage) of aspect terms or opinion words that have been correctly classified. Random assignment is the random guess baseline.}
	\label{figure:brown_clusters_number}
	
\end{figure*}

We perform several experiments varying the number of Brown clusters involved in the process (see section \ref{sec:atow_separation}) to evaluate if it has a noticeable impact on the word separation.
Figure \ref{figure:brown_clusters_number} shows the resulting proportions of correctly assigned aspect terms and opinion words for English. In general, the correct proportions are high compared to a random assignment, which indicates that the aspect-term/opinion-word separation performs correctly most of the times. Interestingly, aspect-terms are better distinguished than opinion-words.


\section{Conclusions and future work}
\label{sec:conclusions}

In this document, we have presented W2VLDA, a system that performs aspect and sentiment classification with almost no supervision and without the need of language or domain specific resources. In order to do that, the system combines different unsupervised approaches, like word embeddings or Latent Dirichlet Allocation (LDA), to bootstrap information from a domain corpus. The only supervision required by the user is a single seed word per desired aspect and polarity. Because of that, the system can be applied to datasets of different languages and domains with almost no adaptation.
The resulting topics and polarities are directly paired with the aspect names selected by the user at the beginning, so the output can be used to perform Aspect Based Sentiment Analysis.
In addition, the system tries to separate automatically aspect terms and opinion words, providing more clear information and insight to the resulting domain aspects vocabulary.
We evaluate W2VLDA for aspect classification using customer reviews of several domains and compare it against other LDA-based approaches. We also evaluate its performance using a subset of the multilingual SemEval 2016 task 5 ABSA dataset.
As future work, it would we interesting to include an automated way to deal with stop-words and other words that do not carry information for the ABSA task. A better-integrated handling of multi-word and negation expressions could also improve the results.
On the other hand, the are more specialised word embeddings related to sentiment analysis \citep{rothe2016ultradense}, and it would be interesting to study if different word embeddings bring improvements to the method keeping a minimal supervision.

 \section*{Acknowledgements}
 
This work was supported by Vicomtech-IK4 and by the project TUNER - TIN2015-65308-C5-1-R (MINECO/FEDER, UE).

\section*{References}





\bibliographystyle{apa}
\bibliography{references}

\begin{thebibliography}{}

\bibitem[\protect\astroncite{Agerri and Rigau}{2016}]{Agerri201663}
Agerri, R. and Rigau, G. (2016).
\newblock Robust multilingual named entity recognition with shallow
  semi-supervised features.
\newblock {\em Artificial Intelligence}, 238:63 -- 82.

\bibitem[\protect\astroncite{Alam et~al.}{2016}]{Alam2016}
Alam, M.~H., Ryu, W.~J., and Lee, S.~K. (2016).
\newblock {Joint multi-grain topic sentiment: Modeling semantic aspects for
  online reviews}.
\newblock {\em Information Sciences}, 339:206--223.

\bibitem[\protect\astroncite{Araque et~al.}{2017}]{araque2017enhancing}
Araque, O., Corcuera-Platas, I., S{\'a}nchez-Rada, J.~F., and Iglesias, C.~A.
  (2017).
\newblock Enhancing deep learning sentiment analysis with ensemble techniques
  in social applications.
\newblock {\em Expert Systems with Applications}, 77:236--246.

\bibitem[\protect\astroncite{Bhatia et~al.}{2016}]{bhatia2016automatic}
Bhatia, S., Lau, J.~H., and Baldwin, T. (2016).
\newblock Automatic labelling of topics with neural embeddings.
\newblock {\em arXiv preprint arXiv:1612.05340}.

\bibitem[\protect\astroncite{Blair-Goldensohn
  et~al.}{2008}]{Blair-Goldensohn2008}
Blair-Goldensohn, S., Hannan, K., McDonald, R., Neylon, T., Reis, G.~A., and
  Reynar, J. (2008).
\newblock Building a sentiment summarizer for local service reviews.
\newblock In {\em WWW workshop on NLP in the information explosion era},
  volume~14, pages 339--348.

\bibitem[\protect\astroncite{Blei et~al.}{2003}]{blei2003latent}
Blei, D.~M., Ng, A.~Y., and Jordan, M.~I. (2003).
\newblock Latent dirichlet allocation.
\newblock {\em the Journal of machine Learning research}, 3:993--1022.

\bibitem[\protect\astroncite{Brody and Elhadad}{2010}]{Brody2010}
Brody, S. and Elhadad, N. (2010).
\newblock {An unsupervised aspect-sentiment model for online reviews}.
\newblock {\em The 2010 Annual Conference of the North American Chapter of the
  Association for Computational Linguistics}, (June):804--812.

\bibitem[\protect\astroncite{Brown et~al.}{1992}]{Brown1992}
Brown, P.~F., Desouza, P.~V., Mercer, R.~L., Pietra, V. J.~D., and Lai, J.~C.
  (1992).
\newblock Class-based n-gram models of natural language.
\newblock {\em Computational linguistics}, 18(4):467--479.

\bibitem[\protect\astroncite{Chen et~al.}{2017}]{chen2017improving}
Chen, T., Xu, R., He, Y., and Wang, X. (2017).
\newblock {Improving sentiment analysis via sentence type classification using
  BiLSTM-CRF and CNN}.
\newblock {\em Expert Systems with Applications}, 72:221--230.

\bibitem[\protect\astroncite{Chen et~al.}{2014}]{Liu2014a}
Chen, Z., Mukherjee, A., and Liu, B. (2014).
\newblock {Aspect extraction with automated prior knowledge learning}.
\newblock {\em Proceedings of the 52nd Annual Meeting of the Association for
  Computational Linguistics}, pages 347--358.

\bibitem[\protect\astroncite{Collobert and Weston}{2008}]{collobert2008unified}
Collobert, R. and Weston, J. (2008).
\newblock A unified architecture for natural language processing: Deep neural
  networks with multitask learning.
\newblock In {\em Proceedings of the 25th international conference on Machine
  learning}, pages 160--167. ACM.

\bibitem[\protect\astroncite{Das et~al.}{2015}]{das2015gaussian}
Das, R., Zaheer, M., and Dyer, C. (2015).
\newblock {Gaussian LDA for Topic Models with Word Embeddings.}
\newblock In {\em Proceedings of the 53nd Annual Meeting of the Association for
  Computational Linguistics}, pages 795--804.

\bibitem[\protect\astroncite{Ganu et~al.}{2009}]{Ganu2009}
Ganu, G., Elhadad, N., and Marian, A. (2009).
\newblock Beyond the stars: Improving rating predictions using review text
  content.
\newblock In {\em WebDB}, volume~9, pages 1--6. Citeseer.

\bibitem[\protect\astroncite{Giatsoglou et~al.}{2017}]{giatsoglou2017sentiment}
Giatsoglou, M., Vozalis, M.~G., Diamantaras, K., Vakali, A., Sarigiannidis, G.,
  and Chatzisavvas, K.~C. (2017).
\newblock Sentiment analysis leveraging emotions and word embeddings.
\newblock {\em Expert Systems with Applications}, 69:214--224.

\bibitem[\protect\astroncite{Griffiths and
  Steyvers}{2004}]{griffiths2004finding}
Griffiths, T.~L. and Steyvers, M. (2004).
\newblock Finding scientific topics.
\newblock {\em Proceedings of the National Academy of Sciences}, 101(suppl
  1):5228--5235.

\bibitem[\protect\astroncite{Hu and Liu}{2004}]{Hu2004}
Hu, M. and Liu, B. (2004).
\newblock Mining opinion features in customer reviews.
\newblock In {\em AAAI}, volume~4, pages 755--760.

\bibitem[\protect\astroncite{Huang et~al.}{2014}]{huang2014automatic}
Huang, S., Niu, Z., and Shi, C. (2014).
\newblock Automatic construction of domain-specific sentiment lexicon based on
  constrained label propagation.
\newblock {\em Knowledge-Based Systems}, 56:191--200.

\bibitem[\protect\astroncite{Jijkoun et~al.}{2010}]{jijkoun2010generating}
Jijkoun, V., de~Rijke, M., and Weerkamp, W. (2010).
\newblock Generating focused topic-specific sentiment lexicons.
\newblock In {\em Proceedings of the 48th Annual Meeting of the Association for
  Computational Linguistics}, pages 585--594. Association for Computational
  Linguistics.

\bibitem[\protect\astroncite{Jo and Oh}{2011}]{Jo2011}
Jo, Y. and Oh, A.~H. (2011).
\newblock Aspect and sentiment unification model for online review analysis.
\newblock In {\em Proceedings of the fourth ACM international conference on Web
  search and data mining}, pages 815--824. ACM.

\bibitem[\protect\astroncite{Kim et~al.}{2013}]{KimSuin2013}
Kim, S., Zhang, J., Chen, Z., Oh, A., and Liu, S. (2013).
\newblock {A Hierarchical Aspect-Sentiment Model for Online Reviews.}
\newblock {\em Proceedings of the Twenty-Seventh AAAI Conference on Artificial
  Intelligence}, pages 526--533.

\bibitem[\protect\astroncite{Lin et~al.}{2011}]{Lin2012}
Lin, C., He, Y., Everson, R., and R{\"{u}}ger, S. (2011).
\newblock {Weakly supervised joint sentiment-topic detection from text}.
\newblock {\em IEEE Transactions on Knowledge and Data Engineering},
  24:1134--1145.

\bibitem[\protect\astroncite{Lin et~al.}{2009}]{Lin2009}
Lin, C., Road, N.~P., and Ex, E. (2009).
\newblock {Joint Sentiment / Topic Model for Sentiment Analysis}.
\newblock {\em Cikm}, pages 375--384.

\bibitem[\protect\astroncite{Liu}{2012}]{liu2012sentiment}
Liu, B. (2012).
\newblock Sentiment analysis and opinion mining.
\newblock {\em Synthesis Lectures on Human Language Technologies}, 5(1):1--167.

\bibitem[\protect\astroncite{Liu et~al.}{2012}]{Liu2012}
Liu, K., Xu, L., and Zhao, J. (2012).
\newblock {Opinion target extraction using word-based translation model}.
\newblock {\em Proceedings of the 2012 Joint Conference on Empirical Methods in
  Natural Language Processing and Computational Natural Language Learning},
  (July):1346--1356.

\bibitem[\protect\astroncite{Lu et~al.}{2011}]{Lu2011}
Lu, B., Ott, M., Cardie, C., and Tsou, B.~K. (2011).
\newblock {Multi-aspect sentiment analysis with topic models}.
\newblock {\em Proceedings - IEEE International Conference on Data Mining,
  ICDM}, pages 81--88.

\bibitem[\protect\astroncite{Mcauliffe and
  Blei}{2008}]{mcauliffe2008supervised}
Mcauliffe, J.~D. and Blei, D.~M. (2008).
\newblock Supervised topic models.
\newblock In {\em Advances in neural information processing systems}, pages
  121--128.

\bibitem[\protect\astroncite{Mikolov et~al.}{2013a}]{Mikolov2013}
Mikolov, T., Chen, K., Corrado, G., and Dean, J. (2013a).
\newblock {Efficient Estimation of Word Representations in Vector Space}.
\newblock {\em arXiv preprint arXiv:1301.3781}, pages 1--12.

\bibitem[\protect\astroncite{Mikolov et~al.}{2013b}]{Mikolov2013lingRegu}
Mikolov, T., Yih, W.-t., and Zweig, G. (2013b).
\newblock {Linguistic regularities in continuous space word representations}.
\newblock {\em Proceedings of NAACL-HLT}, pages 746--751.

\bibitem[\protect\astroncite{Mostafa}{2013}]{mostafa2013more}
Mostafa, M.~M. (2013).
\newblock More than words: Social networks’ text mining for consumer brand
  sentiments.
\newblock {\em Expert Systems with Applications}, 40(10):4241--4251.

\bibitem[\protect\astroncite{Mukherjee and Liu}{2012}]{Mukherjee2012}
Mukherjee, A. and Liu, B. (2012).
\newblock {Aspect extraction through semi-supervised modeling}.
\newblock {\em ACL '12 Proceedings of the 50th Annual Meeting of the
  Association for Computational Linguistics: Long Papers - Volume 1},
  (July):339--348.

\bibitem[\protect\astroncite{Nguyen et~al.}{2015}]{Nguyen2015}
Nguyen, D.~Q., Billingsley, R., Du, L., and Johnson, M. (2015).
\newblock {Improving topic models with latent feature word representations}.
\newblock {\em Transactions of the Association for Computational Linguistics},
  3:299--313.

\bibitem[\protect\astroncite{Pang and Lee}{2008}]{pang2008opinion}
Pang, B. and Lee, L. (2008).
\newblock Opinion mining and sentiment analysis.
\newblock {\em Foundations and trends in information retrieval}, 2(1-2):1--135.

\bibitem[\protect\astroncite{Pontiki et~al.}{2016}]{pontiki-EtAl:2016:SemEval}
Pontiki, M., Galanis, D., Papageorgiou, H., Androutsopoulos, I., Manandhar, S.,
  AL-Smadi, M., Al-Ayyoub, M., Zhao, Y., Qin, B., De~Clercq, O., Hoste, V.,
  Apidianaki, M., Tannier, X., Loukachevitch, N., Kotelnikov, E., Bel, N.,
  Jim\'{e}nez-Zafra, S.~M., and Eryi\u{g}it, G. (2016).
\newblock Semeval-2016 task 5: Aspect based sentiment analysis.
\newblock In {\em Proceedings of the 10th International Workshop on Semantic
  Evaluation (SemEval-2016)}, pages 19--30, San Diego, California. Association
  for Computational Linguistics.

\bibitem[\protect\astroncite{Pontiki et~al.}{2015}]{pontiki2015semeval}
Pontiki, M., Galanis, D., Papageorgiou, H., Manandhar, S., and Androutsopoulos,
  I. (2015).
\newblock Semeval-2015 task 12: Aspect based sentiment analysis.
\newblock In {\em Proceedings of the 9th International Workshop on Semantic
  Evaluation (SemEval 2015), Association for Computational Linguistics, Denver,
  Colorado}, pages 486--495.

\bibitem[\protect\astroncite{Pontiki et~al.}{2014}]{Pontiki2014}
Pontiki, M., Galanis, D., Pavlopoulos, J., Papageorgiou, H., Androutsopoulos,
  I., and Manandhar, S. (2014).
\newblock Semeval-2014 task 4: Aspect based sentiment analysis.
\newblock {\em Proceedings of the 8th International Workshop on Semantic
  Evaluation (SemEval 2014), Association for Computational Linguistics, Dublin,
  Ireland}, pages 27--35.

\bibitem[\protect\astroncite{Popescu and Etzioni}{2007}]{Popescu2005}
Popescu, A.-M. and Etzioni, O. (2007).
\newblock Extracting product features and opinions from reviews.
\newblock In {\em Natural language processing and text mining}, pages 9--28.
  Springer.

\bibitem[\protect\astroncite{Qiang et~al.}{2016}]{Qiang2016}
Qiang, J., Chen, P., Wang, T., and Wu, X. (2016).
\newblock {Topic Modeling over Short Texts by Incorporating Word Embeddings}.
\newblock {\em arXiv preprint arXiv: 1609.08496v1}, page~10.

\bibitem[\protect\astroncite{Qiu et~al.}{2011}]{Qiu2011}
Qiu, G., Liu, B., Bu, J., and Chen, C. (2011).
\newblock Opinion word expansion and target extraction through double
  propagation.
\newblock {\em Computational linguistics}, 37(1):9--27.

\bibitem[\protect\astroncite{Ramage et~al.}{2009}]{ramage2009labeled}
Ramage, D., Hall, D., Nallapati, R., and Manning, C.~D. (2009).
\newblock {Labeled LDA: A supervised topic model for credit attribution in
  multi-labeled corpora}.
\newblock In {\em Proceedings of the 2009 Conference on Empirical Methods in
  Natural Language Processing: Volume 1-Volume 1}, pages 248--256. Association
  for Computational Linguistics.

\bibitem[\protect\astroncite{Rao and Ravichandran}{2009}]{rao2009semi}
Rao, D. and Ravichandran, D. (2009).
\newblock Semi-supervised polarity lexicon induction.
\newblock In {\em Proceedings of the 12th Conference of the European Chapter of
  the Association for Computational Linguistics}, pages 675--682. Association
  for Computational Linguistics.

\bibitem[\protect\astroncite{Rothe et~al.}{2016}]{rothe2016ultradense}
Rothe, S., Ebert, S., and Sch{\"u}tze, H. (2016).
\newblock Ultradense word embeddings by orthogonal transformation.
\newblock {\em arXiv preprint arXiv:1602.07572}.

\bibitem[\protect\astroncite{Tang et~al.}{2014}]{Tang2014}
Tang, D., Wei, F., Yang, N., Zhou, M., Liu, T., and Qin, B. (2014).
\newblock {Learning Sentiment-Specific Word Embedding}.
\newblock {\em Acl}, pages 1555--1565.

\bibitem[\protect\astroncite{Turian et~al.}{2010}]{turian2010word}
Turian, J., Ratinov, L., and Bengio, Y. (2010).
\newblock Word representations: a simple and general method for semi-supervised
  learning.
\newblock In {\em Proceedings of the 48th annual meeting of the association for
  computational linguistics}, pages 384--394. Association for Computational
  Linguistics.

\bibitem[\protect\astroncite{Turney and Littman}{2003}]{turney2003measuring}
Turney, P.~D. and Littman, M.~L. (2003).
\newblock Measuring praise and criticism: Inference of semantic orientation
  from association.
\newblock {\em ACM Transactions on Information Systems (TOIS)}, 21(4):315--346.

\bibitem[\protect\astroncite{Wu et~al.}{2009}]{wu2009phrase}
Wu, Y., Zhang, Q., Huang, X., and Wu, L. (2009).
\newblock Phrase dependency parsing for opinion mining.
\newblock In {\em Proceedings of the 2009 Conference on Empirical Methods in
  Natural Language Processing}, volume~3, pages 1533--1541. Association for
  Computational Linguistics.

\bibitem[\protect\astroncite{Zhao et~al.}{2010}]{Zhao2010}
Zhao, W.~X., Jiang, J., Yan, H., and Li, X. (2010).
\newblock {Jointly Modeling Aspects and Opinions with a MaxEnt-LDA Hybrid}.
\newblock {\em Computational Linguistics}, 16(October):56--65.

\end{thebibliography}







\end{document}